\documentclass{tlp}
\usepackage{aopmath}

\usepackage{amsfonts}
\usepackage{url}
\usepackage{epsfig}

\newcommand{\freccia}{\mapsto}

\newcommand{\aminoset}{\mathbb{A}}
\newcommand{\calpha}{C\hspace*{-.25ex}\alpha}

\begin{document}
\bibliographystyle{acmtrans}

\title{CLP-based protein fragment assembly}

\author[Dal Pal\`{u}, Dovier, Fogolari, Pontelli]{
\begin{tabular}{ccc}
Alessandro Dal Pal\`{u}   &\phantom{aaa}& Agostino Dovier \\
Univ. Parma && Univ. Udine\\
Dept. Mathematics & & Dept. Maths and Computer Science\\
\phantom{aaa}\\
Federico Fogolari &\phantom{aaa}& Enrico Pontelli \\
Univ. Udine && New Mexico State Univ.\\
Dept. Biomedical Sciences & & Dept. Computer Science
\end{tabular}
}

\pagerange{\pageref{firstpage}--\pageref{lastpage}}
\volume{\textbf{10} (3):}
\jdate{February 2010}
\setcounter{page}{709}
\pubyear{2010}

\submitted{7 February 2010}
\revised{29 April 2010}
\accepted{12 May 2010}

\label{firstpage}

\maketitle

{\bf Note:} This article has been published in Theory and Practice of Logic Programming,
{\bf 10} (4-6): 709–-724, July
\copyright Cambridge University Press.

\begin{abstract}
The paper investigates a novel approach, based on Constraint Logic
Programming (CLP), to predict the 3D conformation of a protein via
fragments assembly. The
fragments are extracted by a preprocessor---also developed for this
work---
from a database of known protein structures that clusters
and classifies the fragments according to similarity and
frequency. The problem of assembling fragments into a complete
conformation is mapped to a constraint solving problem and solved
using CLP. The constraint-based model uses a
medium discretization degree 
$\calpha$-side chain
centroid protein model that
offers efficiency and a good approximation for space filling.
The approach 
adapts existing energy models to the protein representation used and applies a large neighboring search strategy.
The
results shows the feasibility and efficiency of the
method. The declarative nature of the solution allows to include future extensions,
e.g.
different size fragments for better accuracy.
\end{abstract}

\section{Introduction}
Proteins are central components in the way they control and execute
the vital functions in living organisms. The functions of a protein
are directly
related to
its peculiar three-dimensional conformation.
Knowledge of the three-dimensional conformation of a protein (also
known as the \emph{native conformation} or \emph{tertiary structure})
is essential to biomedical investigation. The native conformation
represents the functional protein and determines how it can interact
with other proteins and affect the functions of the
hosting organism. It is impossible to clearly understand the behavior and
phenotype of an organism without knowledge of the native
conformation of the proteins coded in its genome.
As a result of advancements in DNA sequencing techniques,
there is a large and growing number of protein
sequences---i.e., lists of amino acids,
also known as \emph{primary structures} of proteins---available
 in public databases (e.g., the database Swiss-prot contains about $500,000$
protein sequences). On the other hand, knowledge
of structural information  (e.g.,
information concerning the tertiary structures) is lagging behind,
with a much smaller number of structures deposited in public
databases, notwithstanding that structural genomics initiatives started
worldwide.

For these reasons, one of the most traditional and
central problems addressed by research in bioinformatics deals with
the \emph{protein structure prediction} problem, i.e.,  the problem of
using computational methods to
determine the native conformation of a protein starting from its
primary sequence.
Several approaches have been explored to address this problem.
In broad terms, it is possible to distinguish between two main
classes of approaches.
The more traditional term \emph{protein structure prediction} has been commonly
used to describe methods that rely on comparisons between known and unknown
proteins to predict
the end-result of the spontaneous protein folding process (also known as \emph{homology modeling}). The more expensive task of predicting a protein structure starting from the knowledge of the chemical structure and laws of physics (known as \emph{de-novo/ab-initio prediction}) has been studied with different levels of accuracy and complexity.

Instead, \emph{protein folding} simulations tries to
understand the folding path leading to the native conformation, typically using investigations
of the potential energy landscape or using molecular dynamics simulations. In both classes
of methods, knowledge of ``patterns'' can be used to restrict the search space---and this
is particularly true for the case of \emph{secondary structure} components of a
protein, i.e., local helices or strands.
Secondary structure components are important, considering that their formation is believed to represent
the earliest phase of the folding process, and their identification can be relatively
simpler (e.g., through low-resolution observations of images from electron microscopy).

\subsection{Related work}

As mentioned above, it is possible to predict
protein structures based on their sequences, using either
homology modeling or fold recognition techniques. Nevertheless,
it is in general
difficult to predict a protein structure based {\sl only} on its
sequence and in absence of structural templates.
Explicit solvent
molecular dynamics simulations of protein folding are still beyond
current computational capabilities.
Already in 1968, Levinthal postulated that the systematic
exploration of the space of possible conformations is
infeasible~\cite{leventhal}. This complexity has been confirmed by
theoretical results, showing that even extremely simplified
formalizations of the problem lead to computationally intractable
problems~\cite{crescenzi}. Recently, ab-initio methods for generating
protein structures given their sequences have been proposed and
successfully used~\cite{CASP8}. Key elements of these methods are
the use of evolutionary information from multiple alignments,
the adoption of
simplified representations of proteins, and the introduction
of fragments assembly techniques. These methods rely on assembling the structure using
simplified representations of protein fragments  with favorable conformations
(obtained from structural
databases) for the profile of the given
sequence. Three to nineteen-residue fragments
(i.e., 3--9 for small proteins and 5--19 for
large ones) contain correlations
among neighboring residue conformations, and most of the
computation time is spent in  finding the global arrangement of
fragments that already display good local
conformations~\cite{Rosetta09,Tasser}.

Several simplified models have been introduced to address the
problem. Simplified models abstract several properties of both proteins
and space, leading to a version of the problem that can be solved
more efficiently. The solutions of the simplified problem
constitute candidate configurations that can be refined with more
computationally intensive methods, e.g., molecular dynamics
simulations. Possible simplifications include the introduction of
 fixed sizes of
monomers and bond lengths, the representation of monomers as simple points,
and viewing the three-dimensional space as a discretized collection
of points. In these simplified models, it is possible to view the
protein folding problem as an \emph{optimization problem}, aimed at
determining conformations that minimize an energy function. The
energy function must be defined according to the simplified model
adopted \cite{FEDETAB,BMC07}. Simplified energy models
have been devised specifically to solve large instances of the
structure prediction problem using  constraint solving
approaches~\cite{BacWil06,PASCAL_08}.

In our own previous efforts, we have applied declarative programming
approaches, based on constraint solving, to address
the problem. We built our efforts on a \emph{discrete crystal lattice}
organization of
the allowable
points in the three-dimensional space. This representation exploits the property
that the distance between the $\calpha$ atoms\footnote{A carbon atom that is
a convenient representative of the whole amino acid} of two consecutive
amino acids is relatively constant (3.8\AA). The problem is
viewed as placing amino acids in the allowable points, in such a way
that constraints encoding the mutual distances of amino acids in the
primary sequence are satisfied~\cite{BMC04,SPE07}. The original
framework has also been expanded to support several types of
\emph{global constraints}, i.e., constraints describing complex
relationships among groups of amino acids. One of these constraints
is the \emph{rigid structure} constraint---this constraint enables
the representation of known substructures (e.g., secondary
structure components),
reducing the problem to the determination of an appropriate placement and
rotation of such substructures in the lattice space. The ability to
use rigid structure constraints has been shown to lead to dramatic
reductions in the search space~\cite{IJDMB10,Bara08}. However,
exploiting knowledge of real rigid substructures in a discrete
lattice is infeasible, due to the errors introduced by
the approximations imposed by the
discretized representation of the search space.
On the other hand, the use of a less constrained space model leads to
search spaces that are intractable.
These two considerations lead to the new
approach presented in this work.

\subsection{The contribution of this work}

Some of the most successful approaches to protein folding build on
the principles of using \emph{substructures}. The intuition is that,
while the complete folding of a protein may be unknown, it is likely
that all possible substructures, if properly chosen, can be found
among proteins whose conformations are known.
The folding can be constructed by exploiting
relationships among substructures. A notable example of this approach is
represented by \emph{Rosetta}~\cite{Rosetta09}---an ab initio protein
structure prediction
method that uses simulated annealing search to compose a
conformation, by assembling substructures extracted from a fragment
library; the library is obtained from observed structures stored in the
\emph{Protein Data Bank} (\emph{PDB}, \url{www.pdb.org}).

In this work, we follow a similar idea, by developing a database of
amino acid chains of length $4$; these are clustered according to
similarity, and  their frequencies are drawn from
the investigation of a relevant section of the Protein Data Bank.
The database contains the data needed to solve the protein folding
problem via fragments assembly.
Declarative programming techniques are used to generate clean and compact code,
and to enable rapid prototyping. Moreover, the problem of assembling substructures
is efficiently tackled using the constraint solving techniques provided by $CLP(\mathcal{FD})$
systems.

This paper has the goal of showing that our approach is feasible.
The main advantage, w.r.t.\ a highly engineered and imperative tool, is the modularity of the
constraint system, which offers a convenient framework to test
and integrate statistical data from various predictors and
databases. Moreover, the constrained search technique itself
represents a novel method, compared to popular predictors, and we
show its  effectiveness in combination  with the development  of
new energy functions and heuristics. The proposed solution includes
a general implementation of  large neighboring search in
$CLP(\mathcal{FD})$, that turned
out to be highly effective for the problem at hand.
Another contribution is the development of a new
\emph{energy function} based on two
components: a contact potential for backbone and side chain centroids
interaction, and an energy component for backbone conformational preferences.
Backbone and side chain steric hindrances are imposed as constraints.

\section{Protein Abstraction}

\subsection{Preliminary notions}\label{preliminarysection}

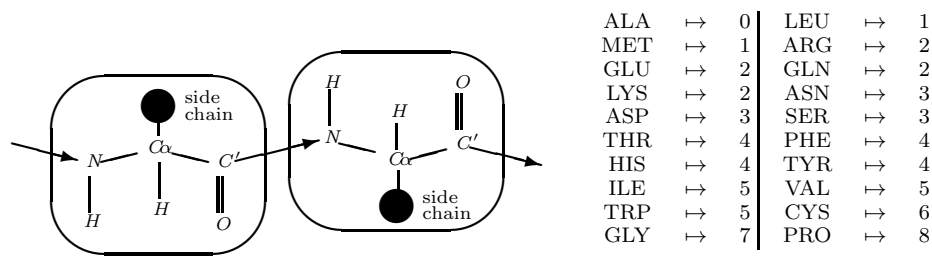
\begin{figure}[ht]
\begin{tabular}{cc}
\scriptsize
\setlength{\unitlength}{0.45pt}
\begin{picture}(210,135)(185,25)
\thicklines
\put(308,90){\oval(180,150)}
\put(308,140){\circle*{30}}
\put(330,142){\scriptsize side}
\put(330,128){\scriptsize chain}
\put(308,115){\line(0,1){20}}
\put(298,100){$C\hspace*{-0.7ex}\alpha$}
\put(309,92){\line(0,-1){30}}
\put(303,48){$H$}
\put(290,100){\line(-4,-1){30}}
\put(247,88){$N$}
\put(185,108){\vector(4,-1){55}}
\put(250,80){\line(0,-1){30}}
\put(245,38){$H$}
\put(320,100){\line(4,-1){30}}
\put(357,88){$C'$}
\put(362,80){\line(0,-1){30}}
\put(358,80){\line(0,-1){30}}
\put(355,35){$O$}
\put(373,93){\vector(4,1){70}}
\thicklines
\put(530,58){\scriptsize side}
\put(530,44){\scriptsize chain}
\put(508,110){\oval(180,150)}
\put(447,108){$N$}
\put(460,108){\line(4,-1){30}}
\put(452,123){\line(0,1){30}}
\put(447,155){$H$}
\put(500,90){$C\hspace*{-0.7ex}\alpha$}
\put(510,85){\line(0,-1){20}}
\put(508,55){\circle*{30}}
\put(508,103){\line(0,1){22}}
\put(503,130){$H$}
\put(520,95){\line(4,1){30}}
\put(557,103){$C'$}
\put(563,120){\line(0,1){30}}
\put(559,120){\line(0,1){30}}
\put(556,155){$O$}
\put(570,105){\vector(4,-1){60}}
\end{picture} &
\footnotesize
\begin{tabular}[b]{ccc|ccc}
ALA & $\freccia$ & 0 &  LEU  & $\freccia$ & 1 \\
MET & $\freccia$ & 1 & ARG & $\freccia$ & 2 \\
GLU  & $\freccia$ & 2 &  GLN & $\freccia$ & 2 \\
LYS & $\freccia$ & 2 &  ASN  & $\freccia$ & 3 \\
ASP & $\freccia$ & 3 & SER  & $\freccia$ & 3 \\
THR & $\freccia$ & 4 & PHE & $\freccia$ & 4 \\
HIS & $\freccia$ & 4 & TYR & $\freccia$ & 4 \\
ILE & $\freccia$ & 5 & VAL & $\freccia$ & 5 \\
TRP & $\freccia$ & 5 & CYS & $\freccia$ & 6 \\
GLY & $\freccia$ & 7 & PRO & $\freccia$ & 8 \\
\end{tabular}
\end{tabular}
\caption{\label{amino_schema}\label{tabella:cluster}\rm
Two consecutive amino acids and the clustering of amino acids into 9 classes}
\end{figure}

\begin{figure}[htbp]
\begin{tabular}{c}
\psfig{figure=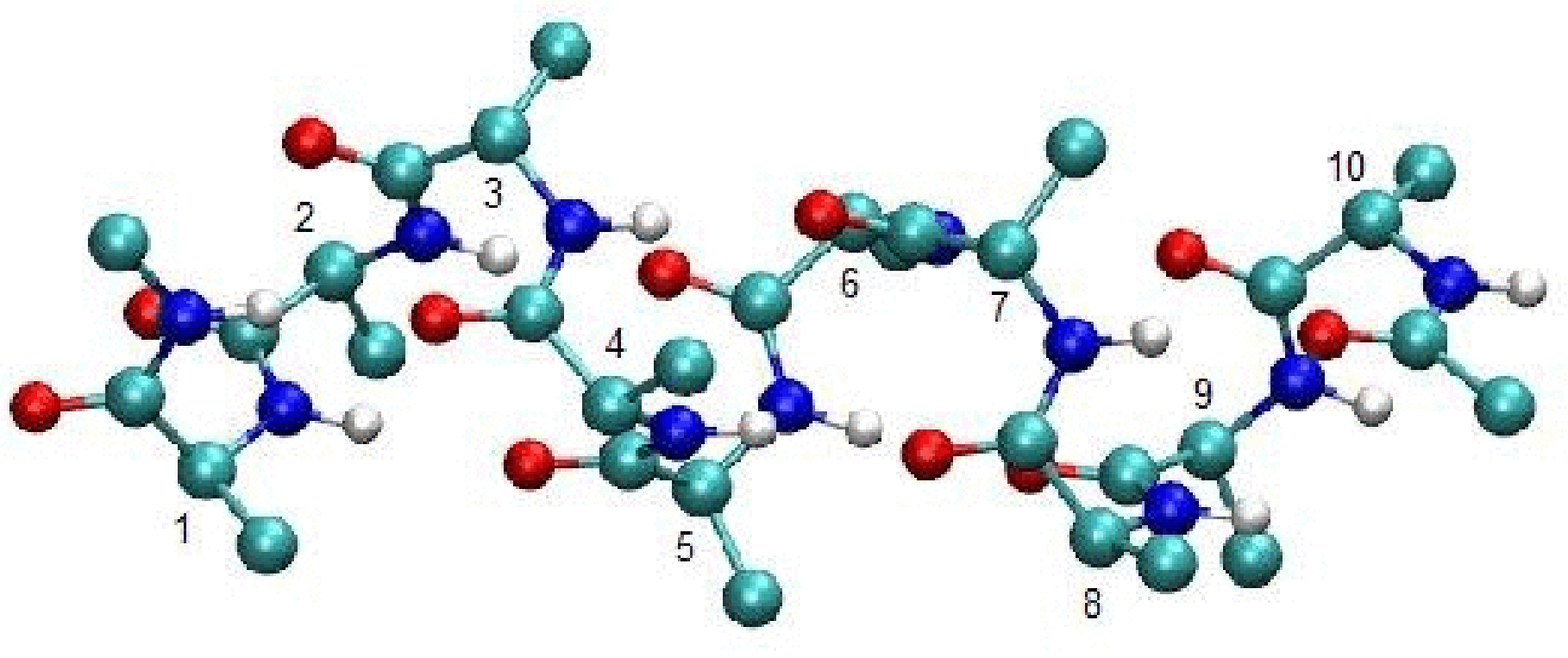,width=0.45\textwidth}
\psfig{figure=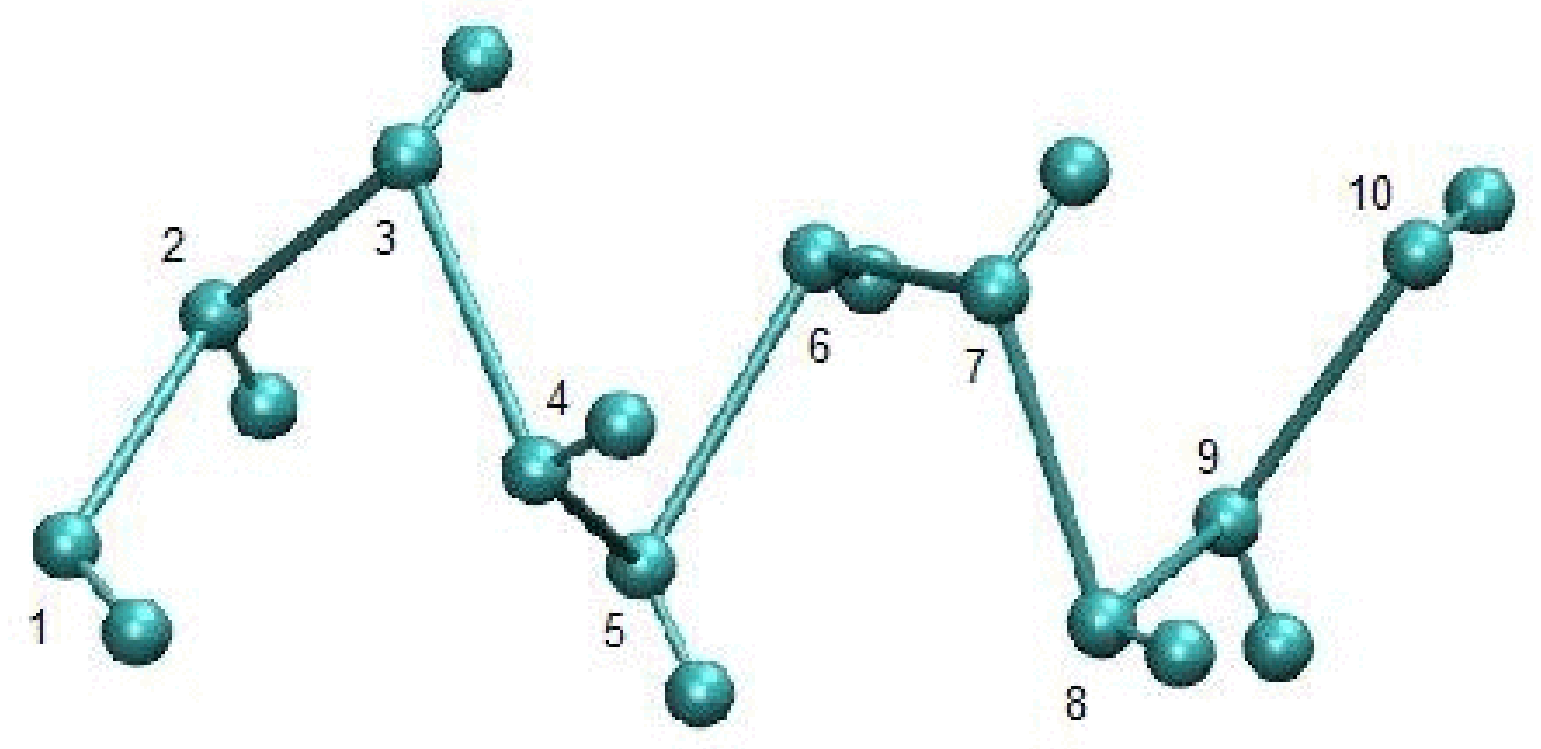,width=0.45\textwidth}
\end{tabular}
\caption{\label{decaala} A fragment of 10 ALA amino acids
in all-atom and $\calpha$-centroid representation}
\end{figure}

We focus on proteins described as sequences of amino acids
selected from a set $\aminoset$ of 20 elements
(those coded by the human genome).
In turn, each amino acid is composed of a set of atoms that
constitute the amino acid's
backbone (see Fig.~\ref{amino_schema}) and a
set of atoms that differentiate pairs of amino acids, known
as \emph{side chain}.
One of the most important structural
properties is that two consecutive $\calpha$ atoms have an average
distance of 3.8\AA.
Side chains may contain from
1 to 18 atoms, depending on the amino acid.
For computational purposes,
instead of considering all atoms composing the protein, we
consider a simplified model in which we are interested in the position
of the $\calpha$ atoms (representing the backbone of the protein)
and of  particular points, known as the \emph{centroids} of the side chains
(Fig. \ref{decaala}). A natural choice for the centroid is the center of mass of the side chain.

It is important to mention that, once the positions of all
the $\calpha$ atoms and of all the centroids are known,
the structure of the protein is already sufficiently determined, i.e.,
 the position of the remaining atoms can be identified
almost deterministically
with a reasonable accuracy.

\begin{figure}[htbp]
\psfig{figure=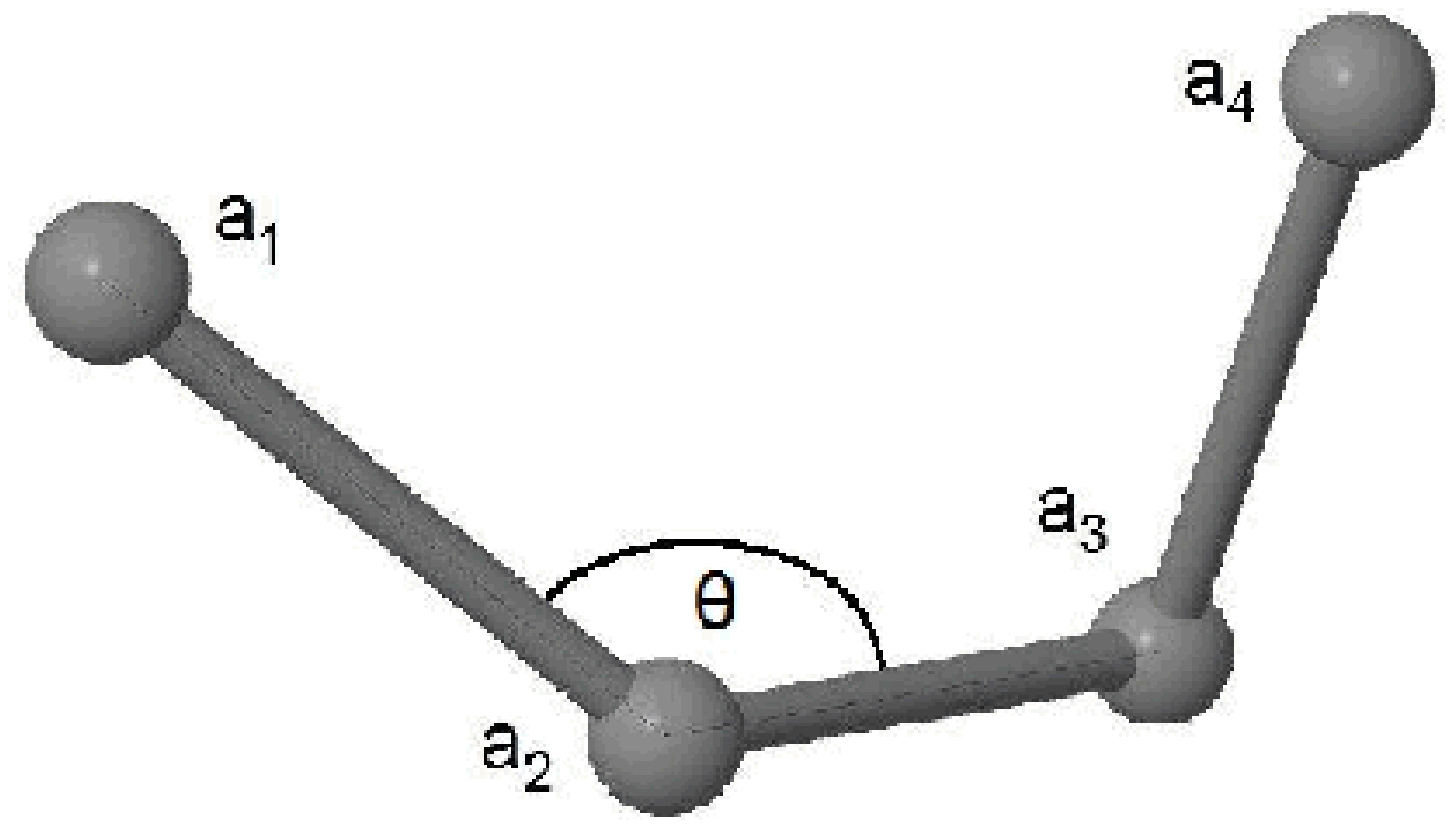,width=0.40\textwidth}
\hspace{.01\textwidth}
\psfig{figure=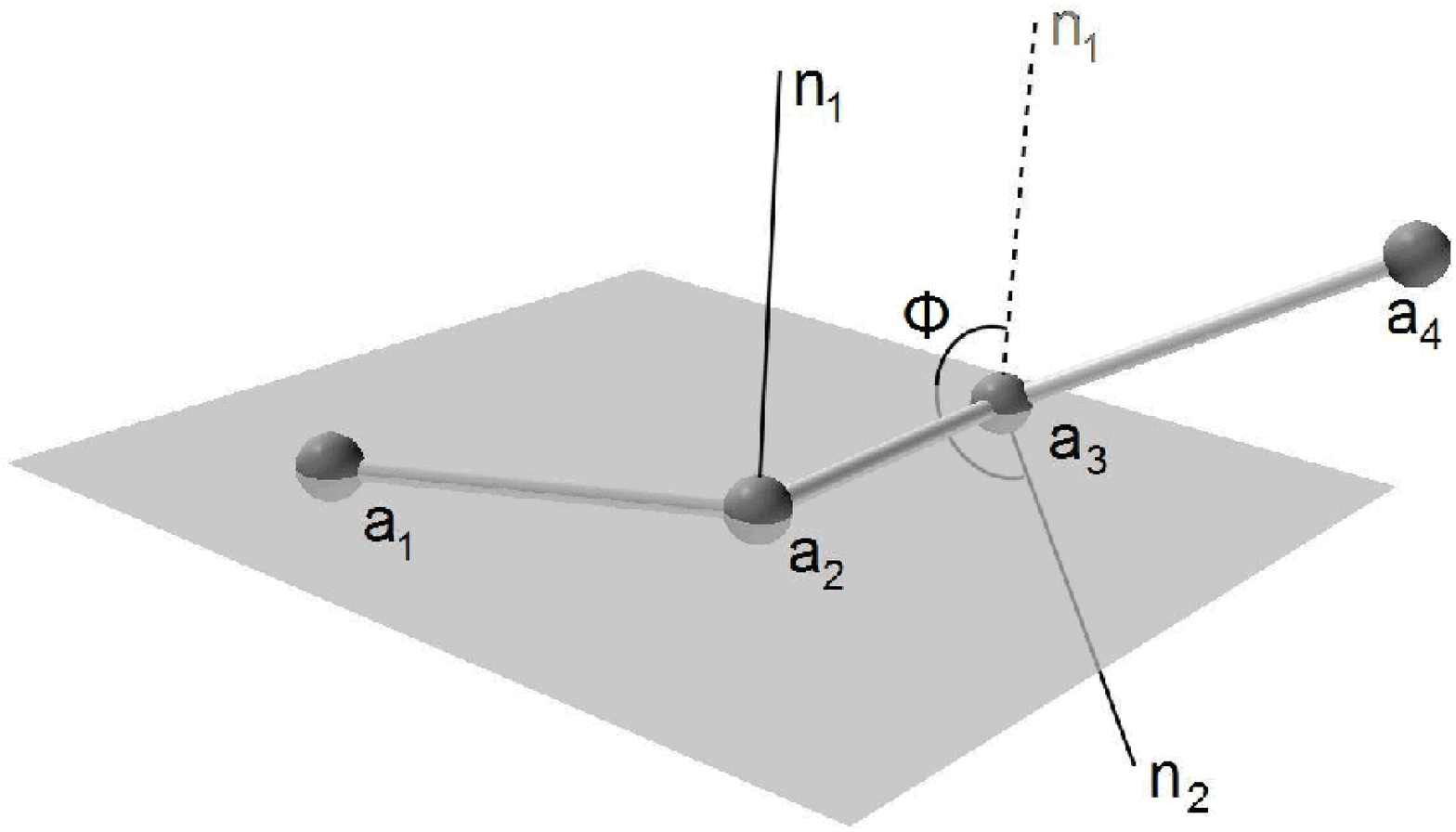,width=0.5\textwidth}
\caption{\label{angle schema}Bend (left) and torsional (right) angles}
\end{figure}

Focusing on the backbone and on the $\calpha$ atoms,
three consecutive amino acids define a \emph{bend} angle
(see $\theta$ in Fig. \ref{angle schema}---left).
Consider now four consecutive amino acids $a_1, a_2, a_3, a_4$.
The angle formed by $n_2 = (a_4-a_3)\times(a_3-a_2)$ and $ n_1 = (a_3-a_2)\times(a_2-a_1)$ is called
\emph{torsional angle}
(see $\phi$ in Fig. \ref{angle schema}---right).
If these angles are known for all the consecutive 4-tuples
forming a protein, they uniquely describe
the 3D positions of all the $\calpha$ atoms of the protein.

Given a spatial conformation of a 4-tuple of consecutive $\calpha$ atoms,
a small degree of freedom for the position of the side chain is
allowed---leading to conformers commonly referred to as
\emph{rotamers}. To reduce the search space,
we do not consider such variations. Once the positions of the $\calpha$ atoms
are known, we  deterministically add the positions of the centroids.
In particular,
the centroid of the $i$-th residue is constructed by using the
positions of $\calpha _{i-1}$, $\calpha _{i}$ and $\calpha _{i+1}$ as reference
and by considering the average of the center of mass of the same amino acid type
centroids, sampled from a non-redundant subset of the PDB. The parameters that
uniquely determine its position are: the average $\calpha$-side chain center
of mass distance, the average bend angle formed by the side chain center of
mass-$\calpha _i$-$\calpha _{i+1}$, and the torsional angle formed by
the $\calpha _{i-1}$-$\calpha _i$-$\calpha _{i+1}$-side chain center of mass.
Even with this simplification, the introduction of
the centroids in the model allows us to better
cope with the layout in the 3D space and to use a richer energy model.
In Fig.~\ref{decaala}, we report an example of this abstraction with
a fragment with 10 alanines (ALA). For these amino acids, the centroids coincide with the only heavy atom of each sidechain.

This has been experimentally shown to produce more accurate results,
without adding extra
complexity w.r.t. a model that considers only
the positions of the $\calpha$ atoms and without the use
of centroids.

\subsection{Clustering}

Although more than 60,000 protein structures are present in the PDB,
 the complete set of known proteins contains too much
redundancy (i.e., very similar proteins deposited in several variants) to
be useful for statistical purposes.
Therefore we focused on a subset
of the PDB called {\sf top-500}~\cite{Lovell03Proteins}.\footnote{Note that
our
program {\tt tuple\_generator}, developed to extract the desired
information, can work on any given set of known proteins.}
This set contains 500 proteins, with $107,138$ occurrences
of amino acids.
The number of different 4-tuples occurring in the set is
precisely $62,831$. Since the number of possible 4-tuples of amino
acids is $|\aminoset^4|={20}^{4} = 160,000$, this means that
most 4-tuples do not appear in the selected set;  even those that
appear, occur too rarely to provide significant statistical information.
For this reason, we decided to cluster amino acids into 9 classes, according to
the similarity of the torsional angles of the pseudo bond between two
consecutive $\calpha$ atoms~\cite{BMC07}.
Note that, in our simplified model, two consecutive $c_\alpha$ atoms do not form a covalent bond and this fact is indicated by the term \emph{pseudo}.

Let $\gamma :\aminoset \longrightarrow \{0,\dots,8\}$ be the function assigning a class
to each amino acid, as defined in Fig.~\ref{tabella:cluster};
for $i \in \{0,\dots,8\}$,
let us denote with $\gamma^{-1}(i) = \{ a \in \aminoset \,:\,
\gamma(a) = i\}$.
In this way,
the majority of the $9^4=6,561$ 4-tuples have a representative in the set
(precisely, there are templates for $5,830$ of them).

A second level of approximation is introduced in deciding when two occurrences of
the same 4-tuple have the ``same'' form. The two tuples are placed
in the same
class when their \emph{Root Mean Square Deviation (RMSD)} is less than or
equal to a given threshold ({\tt
rmsd\_thr}); this threshold is currently set to 1.0\AA. We developed
a C program, called {\tt
tuple\_generator}, that creates  a set of facts of the form:
$$\begin{array}{ll}
{\tt tuple}( & [g_1,g_2,g_3,g_4],\phantom{aa}
              [X^{\alpha}_1, Y^{\alpha}_1, Z^{\alpha}_1,
                X^{\alpha}_2, Y^{\alpha}_2, Z^{\alpha}_2,
                X^{\alpha}_3, Y^{\alpha}_3, Z^{\alpha}_3,
                X^{\alpha}_4, Y^{\alpha}_4, Z^{\alpha}_4],\\
             & \mbox{\tt $g_2$-centroid-list},\phantom{aa}
               \mbox{\tt $g_3$-centroid-list},
              \phantom{aa} {\tt FREQ}, \phantom{aa} {\tt ID}, \phantom{aa} {\tt PID})
\end{array}$$

\noindent where $[g_1,g_2,g_3,g_4] \in \{0,\dots,8\}^4$ identifies the class of each
amino acid, $X^{\alpha}_1,\dots, Z^{\alpha}_4$ are the coordinates
of the 4 $\calpha$ atoms of the 4-tuple,\footnote{Without
loss of generality, we set $(X^{\alpha}_1,Y^{\alpha}_1,Z^{\alpha}_1) = (0,0,0)$.}
${\tt FREQ} \in \{0,\dots,1000\}$ is a frequency factor of the template
w.r.t. all occurrences of the 4-tuple
$g_1,\dots,g_4$ in the set {\sf top-500},
$\tt ID$ is a unique identifier for this fact, and $\tt PID$ is
the first protein found containing this template; this last piece of information
will be printed in the  file we produce as output of the computation, in order to allow one to recover the source of a fragment used for the prediction.

As discussed in Sect.~\ref{preliminarysection}, we model
the position of the centroid of the side chain of every amino acid.
$g_i \in \{0,\dots,8\}$ is a representative of the
class of amino acids $\gamma^{-1}(g_i)$
(e.g., $\gamma^{-1}(2) =
\{ \mathrm{ARG}, \mathrm{GLU},
\mathrm{GLN}, \mathrm{LYS} \}$---see
Fig.~\ref{amino_schema}).
For each amino acid $a \in \gamma^{-1}(g_i)$,
we compute
the position of the centroid corresponding to the positions
$X^{\alpha}_1,\dots, Z^{\alpha}_4$ of the $\calpha$ atoms,
and add it to the {\tt $g_i$-centroid-list}.
Let us observe that we do not add the position of the
first and last centroid in the 4-tuples. As a result, at the end of the
computation, only the centroid
of the first and the last
amino acids of the entire protein will be not set; these
can be assigned using a  straightforward post-processing step.

It is unlikely that a 4-tuple $a_1,\dots,a_4$ that does
not appear in the considered training set will
 occur in
a real protein. Nevertheless, in order to handle these cases, if
$[\gamma(a_1),\dots,\gamma(a_4)]$
has no statistics associated to it,
we map it to the special 4-tuple $[-1,-1,-1,-1]$.
By default, we assign to this unknown tuple the set of
6 most common templates (the number can be easily increased)
among the set of all known templates.
Other special 4-tuples are $[-2,-2,-2,-2]$ and $[-3,-3,-3,-3]$; these
are assigned to $\alpha$-helices and $\beta$-sheets every time a secondary structure constraint is locally enforced on a region of the conformation.
Handling of these special tuples will
be described in detail in Section~\ref{secondary}.

We also introduce an additional collection of facts, based on
the predicate {\tt next}, which are used to relate pairs of
 {\tt tuple} facts. The
relation
${\tt next}({\tt ID}_1,{\tt ID}_2,{\tt Mat})$ holds if the last
three amino acids of the sequence in the {\tt tuple} fact
identified by
${\tt ID}_1$ are identical to the first three amino acids
of the sequence in the {\tt tuple} fact ${\tt ID}_2$,  and
the corresponding $\calpha$ sequences are almost identical---in the
sense that the RMSD between them is at most {\tt rmsd\_thr}.
$\tt Mat$ is the
rotation matrix to align the 1--3 $\calpha$ atoms of the
 4-tuple of ${\tt ID}_2$
with  the  2--4 $\calpha$ atoms of the 4-tuple of ${\tt ID}_1$.

\subsection{Statistical energy}
\label{sec:energy}

The energy function used in this work builds on two components:
\emph{(1)} a contact
potential for side chain and backbone contacts and
\emph{(2)} an energy component for each backbone conformation
based on backbone conformational preferences observed in the database.
The first component uses the table of contact energies described
 in~\cite{FEDETAB}. This set of energies has been shown
  to be accurate, and it is the result of filtering the dataset to allow
  accurate statistical analysis, and a consequent
  fine tuning of the contact definition.
  Since  not all side chain atoms are represented, it is not possible to directly use the
definition of~\cite{FEDETAB}; therefore,
 we retain the table of contact energies of \cite{FEDETAB}, but we adapt the contact
     definition to the side chain centroid. In particular, we use the value
     $3.2$\AA\ as the shortest distance between two centroids, and we set the contact distance
     between two centroids to be less than or equal to the sum of their radii.
 Larger distances provide a contribution that decays quadratically with the distance.

     The radius of each centroid is computed as the centroid's average distance from the
     $\calpha$ atom.
     No contact potential is assigned to the side chain's $\calpha$,
     as the side chain definition already includes the $\calpha$ carbon.
     The energy assigned to the contact between backbone atoms represented by $\calpha$ (not
     included in the side chain definition) is the same energy assigned to ASN-ASN contacts,
     which involve mainly similar chemical groups contacts.
The torsional angle defined by four consecutive $\calpha$ atoms is
    assigned an energy value defined by the potential of the mean force derived by the
    distribution of the corresponding torsional angle in the PDB.
    The procedure has been  thoroughly described in \cite{BMC07}.

\section{Modeling}
In this section, we describe the modeling of the problem of
fragments assembly using constraints over finite domains---specifically,
the type of constraints provided in $CLP(\mathcal{FD})$.
The input is  a list ${\tt Primary}$
of $n$ amino acids. We will denote with $a_i$ the $i^{th}$
element of the primary sequence. We also allow
PDB identifiers as inputs; in this
case, the primary structure of the protein is
retrieved from the PDB.
A list of
 $n-3$ variables ({\tt Code}) is generated.
 The $i$-th variable $C_i$ of {\tt Code} corresponds to the
4-tuple $(\gamma(a_i),\dots, \gamma(a_{i+1}),\gamma(a_{i+2}),
\gamma(a_{i+3}))$.
The possible values for $C_i$ are the $\tt ID$s of the
facts of the form:
$${\tt tuple}([\gamma(a_i),\gamma(a_{i+1}),\gamma(a_{i+2}),
\gamma(a_{i+3})],\_,\_,{\tt Freq},{\tt ID},\_).$$

This set is ordered using the frequency information {\tt Freq}
in decreasing order, and stored in a variable ${\tt ListDom}_i$.

The {\tt next} information is used to impose constraints
between $C_i$ and $C_{i+1}$. Using the combinatorial
constraint {\tt table}, we allow only pairs of
consecutive values supported by the {\tt next} predicate.
Recall that, for each allowed combination of values, the {\tt next} predicate
returns the rotation matrix ${\tt M_{i,i+1}}$, which provides the relative
rotation when the two fragments are best fit.

Another list with $3n$ variables ({\tt Tertiary}) is also generated:
$X^\alpha_i,Y^\alpha_i,Z^\alpha_i$ (resp.,
$X^C_i,Y^C_i,Z^C_i$) denoting the 3D position of the $\calpha$ atoms
(resp., of the centroids).
These variables have integer values (representing a
 precision of $10^{-2}$\AA).

In order to correlate {\tt Code} variables and {\tt Tertiary} variables,
consecutive 4-tuples must be constrained. Let us focus on the
$\calpha$ part; consider two consecutive tuples:
\begin{list}{$\bullet$}{\topsep=1pt \parsep=0pt \itemsep=1pt}
\item  $t_i = a_i, a_{i+1}, a_{i+2}, a_{i+3}$
with code variable $C_i$,  and
\item $t_{i+1} = a_{i+1}, a_{i+2}, a_{i+3}, a_{i+4},$
with code variable $C_{i+1}$ and local coordinates $\vec V_1, \vec V_2, \vec V_3, \vec V_4$.
\end{list}
$t_{i+1}$ is rotated as to best overlap
the points in common with $t_{i}$, and it is placed so that the last point in
$t_{i+1}$ is at 3.8\AA\ from the last point in $t_{i}$.

\begin{figure}[htbp]
\psfig{figure=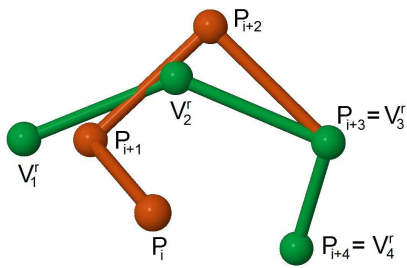,width=0.45\textwidth}
\caption{\label{fig:overlap}Consecutive fragments combination}
\end{figure}

Let $X^{\alpha}_i,Y^{\alpha}_i,Z^{\alpha}_i,\dots,
X^{\alpha}_{i+4},Y^{\alpha}_{i+4},Z^{\alpha}_{i+4}$ be the
variables for the coordinates of these $\calpha$ atoms, stored
in the list {\tt Tertiary} (Fig.~\ref{fig:overlap}, where
$P_i=(X^{\alpha}_i,Y^{\alpha}_i,Z^{\alpha}_i)$).
The constraint introduced rotates and translates the
template $t_{i+1}$ from the reference of $C_i$ (represented by the
orthonormal basis matrix ${\tt R_i}$) according to the rotation matrix
${\tt M_{i,i+1}}$ to the new reference
${\tt R_{i+1}} = {\tt R_i}\times {\tt M_{i,i+1}}$. Moreover, when
placing the template $t_{i+1}$, the constraint affects only the
coordinates of $a_{i+4}$, since the other variables are assigned
by the application of the same constraint for templates
$t_{j}$, $j<i+1$. The constraint shifts the rotated version of $t_{i+1}$ so that
it overlaps the third point $\vec V_3$ with $(X^\alpha_{i+3},Y^\alpha_{i+3},Z^\alpha_{i+3})$.
Formally, let
$\vec V^r_k = {\tt R_{i+1}\times \vec V_k}$, with $k \in \{1 \dots 4\}$, be the
rotated 4-tuple corresponding to $C_{i+1}$. The shift vector
$\vec s=(X^\alpha_{i+3},Y^\alpha_{i+3},Z^\alpha_{i+3})-\vec V^r_3$ is used to
constrain the position of $a_{i+4}$ as follows:

$$(X^\alpha_{i+4},Y^\alpha_{i+4},Z^\alpha_{i+4})=\vec s + {\tt R_{i+1}}\times \vec V_4$$

Note that the 3.8\AA\ distance between consecutive amino
acids (i.e., $a_{i+3}$ and $a_{i+4}$) is preserved, and this constraint allows us
to place templates without requiring an expensive RMSD fit among overlapping
fragments during the search.
Moreover, during a leftmost search, as soon as the variable $C_i$
is assigned, also the coordinates $(X^\alpha_{i+3},Y^\alpha_{i+3},Z^\alpha_{i+3})$
are uniquely determined.

Matrix and vector products are handled by FD variables and constraints,
using a factor of 1000 for storing and handling rotation matrices.

For the sake of simplicity, we omit the formal description of the constraints
associated to the centroids. The centroids' positions
are rotated and shifted accordingly, as soon as the corresponding positions of the $\calpha$ atoms are determined.

The $X_1^\alpha,Y_1^\alpha,Z_1^\alpha,\dots,X_{n}^\alpha,Y_{n}^\alpha,Z_{n}^\alpha$
part of the  {\tt Tertiary} list relative to the
position of the $\calpha$ atoms,
is also required to satisfy a constraint which guarantees
the {\tt all\_distant} property~\cite{IJDMB10}:
the $\calpha$ atoms of each pair of non-consecutive amino acids must be distant
at least $D = 3.2$\AA. This is expressed by the constraint:
$$(X_i^\alpha-X_j^\alpha)^2 +
  (Y_i^\alpha-Y_j^\alpha)^2 +
  (Z_i^\alpha-Z_j^\alpha)^2 \geq D^2$$

\noindent for all $i\in\{1,\dots,n-2\}$ and $j\in\{i+2,\dots,n\}$.
Similar constraints are imposed between pairs of $\calpha$ and
centroids as well as pairs of centroids. In the latter case, in order to account for the differences in volume of
each possible side chain, we determine minimal distances that depend on the
specific type of amino acid considered.

Another constraint is added to guide the search: a {\tt diameter}  parameter is used to bound
the maximum distance between different $\calpha$ atoms (i.e., the diameter of the protein).
As
we argued in earlier work~\cite{BMC04},
a good diameter value is
$5.68 \, n^{0.38}$~\AA.
We impose this constraint
to all different pairs of $\calpha$ atoms.

\subsection{Secondary information}\label{secondary}

The native structure of a protein is largely composed of
some recurrent local structures. These structures,
$\alpha$-helices and $\beta$-sheets, can be predicted with
good accuracy using efficient techniques, such as neural networks,
or recognized using other techniques (e.g., analysis of
density maps from  electron microscopy).
Being based on frequency analysis, our tool is able to discover
the majority of secondary structures. On the other hand,
a-priori knowledge of these structures allows us to
remove several non-deterministic
choices during computation.
Therefore, we allow knowledge of secondary structures
to be provided as part of the input---e.g., information
indicating that the
amino acids $i$--$j$ form an $\alpha$-helix.
In the processing stage, for $k \in \{i,\dots,j-3\}$,
a particular tuple
$[-2,-2,-2,-2]$ is assigned instead of the tuple
$[\gamma(a_k),\dots,\gamma(a_{k+3})]$.
This tuple has a unique template which is associated to
an $\alpha$-helix, built in a standard way using a bend angle of 93.8 degrees
and a torsional angle of 52.3 degrees.
Moreover, a list of the possible positions
for the centroids of the 20 amino acids is retrieved.
Since the domains for
these $C_k$'s are singletons, as soon as $C_i$ is considered for
value assignment, all the points of the helix are deterministically
computed.
The situation in the case of $\beta$-strands is analogous.

A variation of this technique  can be used if larger
and more complex substructures are known.
 Basically, even keeping just 4-tuples as internal data structures,
we can easily deal with tuples of arbitrary lengths.
Automated manipulation of arbitrary complex structures is subject of
future work.

\section{Searching}

The search is guided by the instantiation of the $C_i$
variables. These variables are instantiated in
leftmost-first order; in turn,  the values in their
domains are attempted starting with the most probable value
first.
We observed that a first-fail strategy  does not speed up
the search, probably due to the weak propagation of the
matrix product constraints.
As described in Section~\ref{sec:energy}, we associate an energy to each
computed structure. The energy value is an FD constraint that links
coordinates variables and amino acids. Given the model of the problem, this kind
of constraint is not able to provide effective bounds for pruning the search space
 when searching for optimal solutions. As future work, we plan to investigate specific
 propagators, since the torsional energy contribution could be exploited for
 exact bounds estimations.

Each computed structure is saved in {\tt pdb} format.
This is a standard format for proteins (detailed in the PDB
repository) that can be
processed by most protein viewers
(e.g., Rasmol, ViewerLite, JMol).


In order to further reduce the time to
search for solutions,
we have developed a logic programming
implementation of
\emph{Large Neighboring Search (LNS)}~\cite{Shaw98}.
LNS is a form of local search, where
the search for the successive solutions is performed by exploring
a ``large'' neighborhood. The traditional basic move used in
local search, where the values of a small number of variables
are changed, is replaced by a more general move, where a large
number of variables is allowed to change, and these
variables are subject to constraints.
The basic LNS routine is the following:
\begin{list}{}{\topsep=1pt \parsep=0pt \itemsep=1pt}
\item[1.] Generate an initial solution (e.g., using the standard CLP labeling
    procedure).
\item[2.] Randomly select a subset of
the variables of the problem
that are admissible for changes,
and assign the previous values to the other variables.
\item[3.] Using standard labeling, look for an assignment of the selected variables that
improve the energy/cost function (or look for an
assignment that optimize the
energy/cost function). In any case, go back to step 2.
\end{list}
A timeout mechanism is typically adopted to terminate the cycle
between steps~2 and~3. For example, the procedure is terminated
if either a general time-out occurs or $k$ iterations are performed
without any improvement in the quality of the solution. In these cases,
the best solution found is returned.

This simple procedure can be improved by occasionally allowing
a worsening move---this is important to enable the procedure
to abandon a local minimum. The process can be implemented by
introducing a random variable; whenever such variable is assigned
a certain value (with a low probability), then an alternative move
which worsens the quality of the solution is performed; otherwise we continue using
the scheme described earlier.

The scheme has been implemented in $CLP(\mathcal{FD})$.
Even though the implementation has been developed to meet the needs
of the protein folding problem, the scheme can be adapted with
minimal changes to address the needs of other constraint
optimization problems coded in $CLP(\mathcal{FD})$.
The logical variables of $CLP(\mathcal{FD})$, being single-assignment, are
not suitable to the process of modifying a solution as requested
by LNS. Therefore, a specific combination of backtracking, assertions and reassignment procedures are enforced, in order to reset only the assignments to the CSP (while the constraints are maintained) and to re-assign previous values to the variables that are not going to be changed. The remaining variables can assume any assignment compatible with the constraints.

The implementation we developed avoids these repetitions, by using
extra-logical features of Prolog to record intermediate solutions in
the Prolog database (using the {\tt assert/retract} predicates). The
loss of declarativity is balanced by enhanced performance in the
implementation of LNS.

Fig.~\ref{LNSinprolog} summarizes the $CLP(\mathcal{FD})$ implementation of LNS. The
predicate {\tt best} is used to memorize the best solution encountered, while
the predicate {\tt last\_sol} represents the last solution found. The first
clause of {\tt lns} represents the core of the procedure---by setting up the
constraints ({\tt constraint}) and searching for solutions ({\tt local}); the
{\tt fail} at the end of the clause will cause backtracking into the clauses of
{\tt lns} that determine the final result (lines 7-12). If at least one solution has
been found, then the final result is extracted from the fact of type
{\tt best} in the database.

Starting from one solution (stored in the Prolog database in the fact
{\tt last\_sol}), the predicate {\tt another\_sol} determines the next solution
in the LNS process. If a solution has never been found, then a first solution
is computed, by performing a labeling of the variables (lines 18-19). Otherwise,
the values of some of the variables are modified as dictated by LNS (line 24) and a
new solution is computed. Note that an additional constraint on the resulting value
of the objective function (represented by the variable {\tt Energy}) is added; with
probability $\frac{1}{10}$ (as determined by a random variable {\tt Type} in
line 21) a worsening of the energy is requested (line 22), while in
all other cases an improvement is requested (line 23). If a new solution is found,
this is recorded in the Prolog database (line 32); if this solution is better than
any previous solution, then also the {\tt best} fact is updated (line 36).  Observe
that an internal time-out mechanism (set to 120s---line 25) is applied also on
the search of a new solution.
The ``cut'' in line 30 is introduced to avoid the computation of another solution
with the same random choices.

The iterations of {\tt another\_sol} are performed by the predicate {\tt local}
(lines 13-15). The negated call to {\tt another\_sol} is necessary to enable
removal of all variable assignments (but saving constraints between variables)
each time a new cycle is completed. The {\tt local}
procedure cycles indefinitely.

A final note on the procedure {\tt large\_move}.
We implemented two types of LNS moves. The first
({\tt large\_pivot})
makes a set of consecutive {\tt Code} variables unbound,
allowing the procedure to change a (large) bend. The other
{\tt Code} variables and the first segment of {\tt Tertiary}
variables remain assigned. The second instead leaves unbound
two independent sets of consecutive variables, thus allowing
a rotation of a central part of the protein (a sort of
{\tt large\_crankshaft} move). We use both types of moves
during the search (alternated using a random selection
process).

\begin{figure}
\footnotesize
\begin{verbatim}
1:  lns(ID, Time) :-
2:     constraint(ID,Code,Energy,Primary,Tertiary),
3:     retractall(best(_,_,_)),      assert(best(0,Primary,_)),
4:     retractall(last_sol(_,_,_)),  assert(last_sol(0,null,_)),
5:     time_out(local(Code,Energy,Primary,Tertiary), Time,_),
6:     fail.
7:  lns(_,_) :-
8:     best(0,_,_,_),!,
9:     write('Insufficient time for the first solution'),nl.
10: lns(_,_) :-
11:    best(Energy,Primary,Tertiary),
12:    print_results(Primary,Tertiary,Energy).
13: local(Code,Energy, Primary,Tertiary):-
14:    \+ another_sol(Code,Energy, Primary,Tertiary),
15:    local(Code,Energy, Primary,Tertiary).
16: another_sol(Code, Energy, Primary, Tertiary) :-
17:    last_sol(Lim, LastSol,LastTer),
18:    ( LastSol =  null ->
19:           labeling(Code,Tertiary);
20:      LastSol \= null ->
21:           random(1,11,Type),
22:          (Type =< 1 ->   Lim1 is 5*Lim//6, Energy #> Lim1 ;
23:           Type > 1 ->    Energy #< Lim ),
24:           large_move(Code,LastSol,Tertiary,LastTer),
25:           time_out( labeling(Code,Tertiary), 120000, Flag)),
26:           (Flag == success  -> true ;
27:            Flag == time_out ->
28:               write('Warning: Time out in the labeling stage'),nl,
29:               fail)),
30:     !,
31:     retract(last_sol(_,_,_)),
32:     assert(last_sol(Energy,Code,Tertiary)),
33:     best(Val,_,_),
34:     (Val > Energy ->
35:         retract(best(_,_,_)),
36:         assert(best(Energy,Primary,Tertiary));
37:      true),
38:     fail.
\end{verbatim}
\caption{\label{LNSinprolog}An implementation of LNS in Prolog}
\end{figure}

\section{Experimental results}\label{results}
The prototype  can search the first admissible solution ({\tt
pf\_id(ID,Tertiary)}), where {\tt ID} is a protein name included in
the database ({\tt prot\_list.pl}); the {\tt Primary} sequence is also
admitted directly as input. It is possible to generate
the first {\tt N} solutions and output them as distinct models
in a single pdb file, or to create different files for all the solutions
found within a  {\tt Timeout}.
Finally, LNS can be activated by {\tt pf\_id\_lns(ID,Timeout)}.
A version of the current $CLP(\mathcal{FD})$ implementation,
along with a set of experimental tests,  is
available at
\url{www.dimi.uniud.it/dovier/PF}.

The experimental tests have been performed
on an AMD Opteron 2.2GHz Linux Machine.
Each computation was performed on a single processor using SICStus Prolog 4.0.4.
Some of the experimental results are reported in Table~\ref{tab}.
The execution times reported correspond
to the time needed to compute the best solution
computed within the allowed execution time (s stands for seconds,
m for minutes).

We consider $8$ proteins and perform
an exhaustive search of 6.6 hours.
For other 4 proteins, we perform both enumeration (left) and LNS search for 2 days (right). Moreover we launch LNS for 2 hours (center).

For every protein, we impose the secondary structure
information as specified in the corresponding PDB annotations.
However, we wish to point out that the 12 proteins tested \emph{are not included} in the
{\tt top-500} Database from which we extracted the 4-tuples.
For each protein analyzed using LNS, we performed 4 independent runs,
anticipated by a  {\tt randomize} statement, since the process relies on
random choices. We report the best result (in
terms of RMSD) and the energy associated.

\begin{table}[ht]
\scriptsize\sf
\begin{minipage}{0.49\textwidth}
\renewcommand{\arraystretch}{1.2}
\begin{tabular}{|lr|rrr|}
\cline{1-5}
PID & N & \multicolumn{3}{c|}{Enumerate 6.6 hours}   \\
&& RMSD & \multicolumn{1}{c}{Energy}& \multicolumn{1}{c|}{T (s)}\\
\cline{1-5}
1KVG &12& 2.79& -59122  &9.88 \\
1EDP &17& 3.04& -112755 &73.00 \\
1LE0 &12& 3.12& -45575  & 3.20  \\
    2GP8 &40& 5.96& -266819&5794.88\\
\cline{1-5}
\end{tabular}
\end{minipage}
\begin{minipage}{0.49\textwidth}
\renewcommand{\arraystretch}{1.2}
\begin{tabular}{|lr|rrr|}
\cline{1-5}
PID & N & \multicolumn{3}{c|}{Enumerate 6.6 hours}   \\
&& RMSD & \multicolumn{1}{c}{Energy}& \multicolumn{1}{c|}{T (s)}\\
\cline{1-5}
    1LE3 &16& 3.90& -69017&218.79 \\
    1ENH &54& 5.10& -467014&8553.92  \\
    1PG1 &18& 3.22& -109456 &11.00 \\
    2K9D &44& 6.99& -460877&1453.44\\
\cline{1-5}
\end{tabular}
\end{minipage}
\par\medskip\par
\renewcommand{\arraystretch}{1.2}
\begin{tabular}{|lr|rrr|rrr|rrr|}
\cline{1-11}
PID & N &
\multicolumn{3}{c|}{Enumerate 2 days}
& \multicolumn{3}{c|}{LNS 2 hours}
& \multicolumn{3}{c|}{LNS 2 days}\\
&& RMSD & \multicolumn{1}{c}{Energy}& \multicolumn{1}{c|}{T (m)}
& RMSD &
\multicolumn{1}{c}{Energy}& \multicolumn{1}{c|}{T (m)}
& RMSD &
\multicolumn{1}{c}{Energy}& \multicolumn{1}{c|}{T (m)}
\\
\cline{1-11}
1ZDD & 34& 4.12 & -231469 & 1290  & 3.81& -226387& 6 & 3.81& -226387& 6 \\
1AIL & 69& 9.78 & -711302 & 301  & {5.53}&  -665161&20& {5.44}& -668415 &  109 \\
1VII & 36& 7.06 & -263496 & 1086  & {6.64}& -252231 & 48& {5.52}& -271178 & 170 \\
2IGD & 60& 16.35& -375906 &  2750  & {10.91}& -447513 & 27 &8.67&-467004& 380 \\ 
\cline{1-11}
\end{tabular}

\caption{Computational results}
\label{tab}
\end{table}

The enumeration search is expected to perform better for smaller proteins, where
it is possible to explore a large fraction of the search space within the set
time limit.
It is interesting to note that, for smaller chains, the RMSD w.r.t. the
native conformation in the PDB is rather small (ca. 3\AA); this indicates
that the best solutions found capture the fold of the chain, and the
determined solutions can be refined using molecular dynamics
simulations, as done in~\cite{BMC04}.

Moreover, the proteins 1ZDD, IVII, and 1AIL have been simulated both with enumeration
and LNS search strategy, in order to show that the latter is able to
improve both quality (i.e., RMSD) and computational time
(up to 200 times faster), thanks to the neighborhood exploration.
In the case of 1ZDD and 1AIL, where there are three helices packed together,
the random choice of good pivot moves effectively guides the folding towards
the optimal solution.

Finally, the protein 2IGD shows that the allowed time was not sufficient
 to determine a sensible prediction, even though  the LNS shows enhancements in  quality
 of the prediction, given the same amount of time.
 In this case, additional  work is needed to improve the choice of moves performed
  by  LNS, in order to explore the neighborhood in a more effective way.
  For example, moves that shift and translate some rigid parts of the protein may improve the
  performance and quality.

We also note that the size of the set of fragments in use is sufficient to allow reasonable solutions,
while allowing tractable search spaces. Clearly, the use of a more refined partitioning of the fragments
(by reducing the {\tt rmsd\_thr}) would  produce conformations that are structurally closer to the native state,
at the price of an increase in the size of the search space.

\begin{figure}[ht]
\psfig{figure=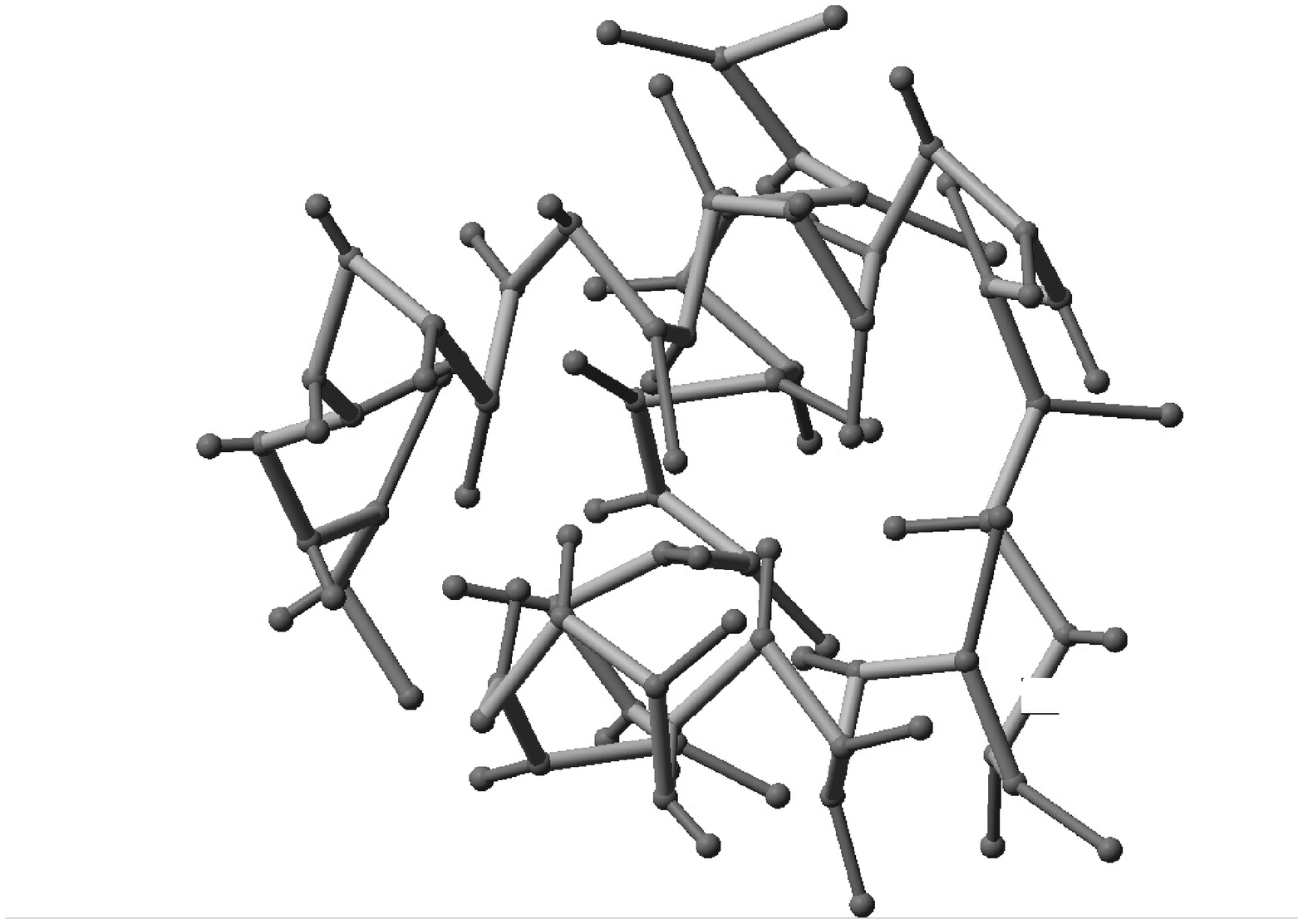,width=0.33\textwidth}
\psfig{figure=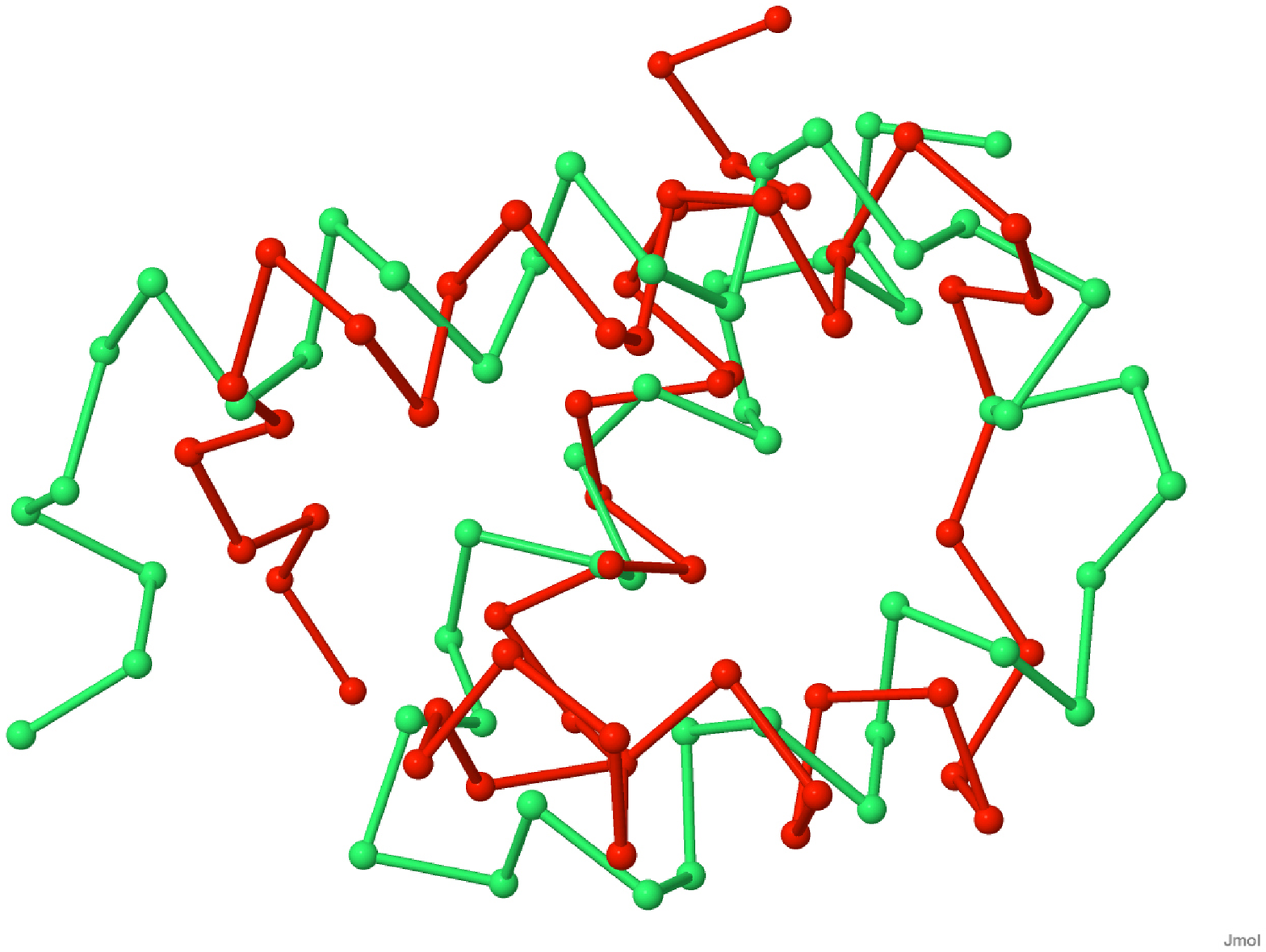,width=0.32\textwidth}
\psfig{figure=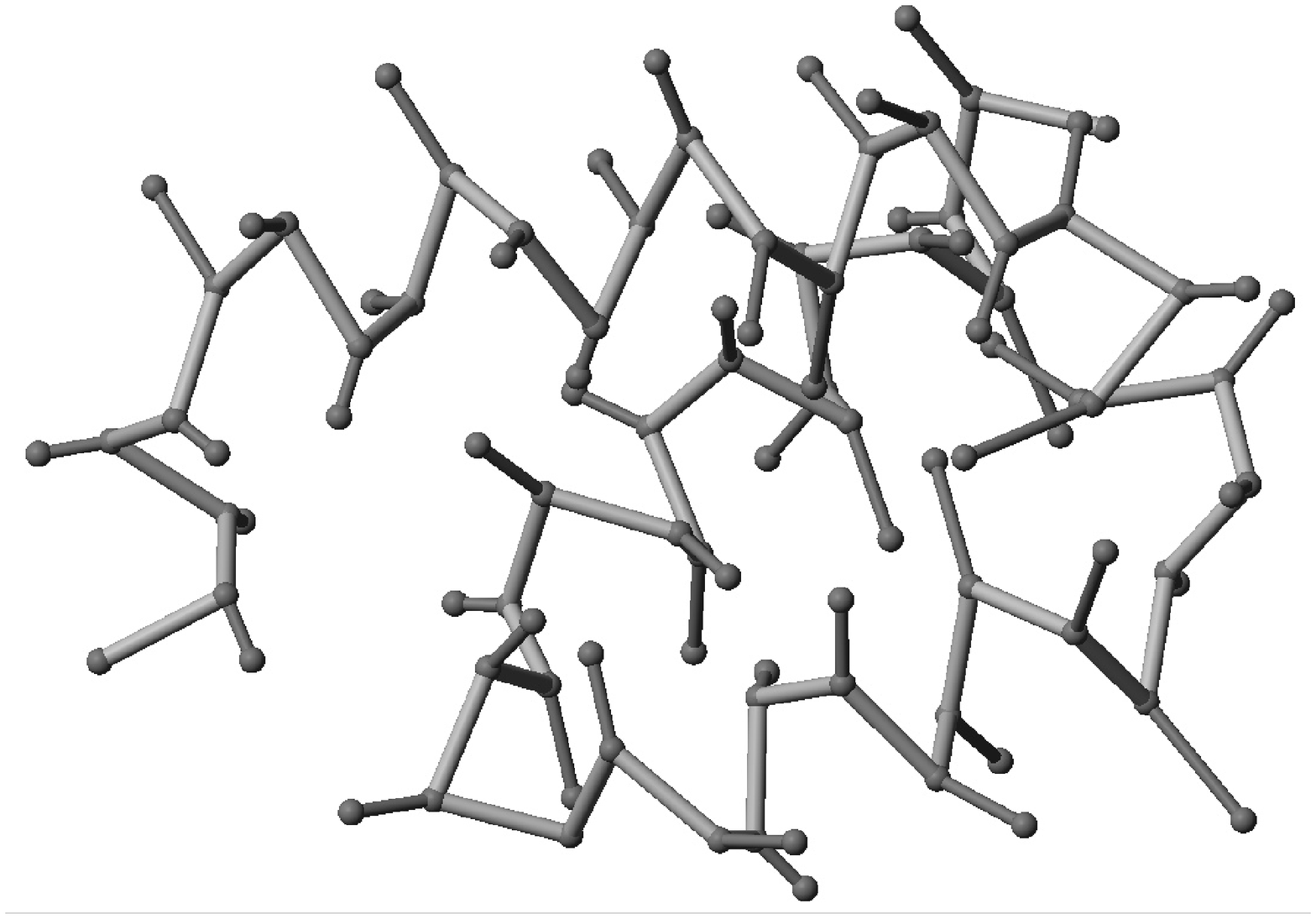,width=0.33\textwidth}
\caption{Protein 1ENH: native state (left), prediction---green/light gray --- compared to original --- red/dark grey--- (center), prediction (right)}
\label{fig:results}
\end{figure}

In Figure~\ref{fig:results}, we depict the native state (on the left) and our
best prediction (on the right) for the protein 1ENH with the backbone and centroids.
In the middle, we show the RMSD overlap of the $C_\alpha$ atoms between the native
conformation (red/dark gray) and the prediction (green/light gray).
The main features are preserved and only the right loop that connects the two helices
appears to have moved significantly.
This could be avoided by introducing a richer set of alternative fragments
in that area and thus allow a more relaxed placement of the fragments.

An important issue is that it is not obvious that the reduced representation and the energy function used here are able to distinguish the native structure from decoys constructed to minimize that energy function. The straightforward comparison between RMSD and energy is not completely meaningful because the native structure should be energy-minimized before energy comparison.
This can be noted by comparing the conformations obtained by
enumeration and LNS. The model in use improves w.r.t. the
$\calpha$ model as presented in~\cite{BMC04}.
However, some fine tuning of the energy coefficients
is still necessary in order to improve the correlation,
while preserving the overall constraint model.
This will be a further area of investigation.
Our results show that the method can scale well and that further
speed-up may be
obtained by considering larger fragments as done by tools like Rosetta.
Rosetta is in fact the state-of-the-art predictor tool (e.g., the small
protein 1ENH is predicted by Rosetta in less than one minute
with a RMSD of 4.2 \AA).

\section{Conclusions and future work}
In this paper we presented the design and implementation of a
constraint logic programming tool to predict the native conformation
of a protein, given its primary structure. The methodology is based
on a process of  fragments assembly, using templates
retrieved from a protein database, that is clustered according to
shape similarity. We used templates based on sequences of
4 amino acids.
The constraint solving process takes advantage of a large
neighboring search strategy, which has been shown to be particularly
effective when dealing with longer proteins.

The preliminary experimental
results confirm the strong potential for this fragment assembly scheme.
The proposed method has a significant advantage over schemes like
Rosetta---the use of $CLP(\mathcal{FD})$ enables the simple addition of ad-hoc
constraints and experimentation with different local search moves
and energy functions.
The implementation presented here constitutes a proof of principle.

In
order to make this a useful tool in a realistic prediction scenario several
improvements must be implemented.
The choice of 4-residue fragment will be improved in the next future
in two directions: fragments will be chosen based on sequence or
profile
alignment (rather than exact match) against a non-redundant
representative set of sequences whose structure is known; the size of
the fragment will be chosen based on the alignment
and will not be restricted to 4-residues.

The reduced representation used here should be replaced by an all-atom
representation at least for the backbone atoms whose hydrogen bonds
define the proper relative arrangement of beta-strands not belonging to
the same fragment. In general we plan to test different
energy functions that better correlate with
RMSD w.r.t.\ the (known) native structures and
the computed ones.
It is likely that with sequences longer than those considered here
predictions will not be equally good in all parts of the molecule,
therefore alternative measurements of similarity like GDT-TS~\cite{Zemla} might
be more appropriate.

Other (possibly redundant)
constraints and constraint propagation techniques should be analyzed,
including the migration to the C++ solver Gecode.

\paragraph{{\bf Acknowledgments.}}

The work is partially supported by the grants:
GNCS-INdAM \emph{Tecniche innovative per la programmazione con vincoli
 in applicazioni strategiche},
PRIN 2007
\emph{Computer simulations of Glutathione Peroxidases: insights
into the catalytic mechanism},
PRIN 2008
\emph{Innovative multi-disciplinary approaches
for constraint and preference reasoning},
NSF HRD-0420407 and NSF IIS-0812267.

\bibliography{iclp_camera}

\end{document}